\def\year{2018}\relax
\documentclass[letterpaper]{article} 
\usepackage{aaai18}  
\usepackage{times}  
\usepackage{helvet}  
\usepackage{courier}  
\usepackage{url}  
\usepackage{graphicx}  

\usepackage{amsfonts}

\def\b{{\bf b}}
\def\r{{\bf r}}

\def\h{{\bf h}}
\def\x{{\bf x}}

\def\z{{\bf z}}

\def\I{{\bf I}}

\def\U{{\bf U}}

\def\W{{\bf W}}

\nocopyright

\title{Gated Orthogonal Recurrent Units: On Learning to Forget}
\author{
Li Jing$^{1*}$, Caglar Gulcehre$^{2*}$, John Peurifoy$^1$, Yichen Shen$^1$, \\
{ \bf \Large
  Max Tegmark$^1$,  Marin Solja\v{c}i\'{c}$^1$, Yoshua Bengio$^2$}\\
  $^1$ Massachusetts Institute of Technology,  
  $^2$ MILA - Universite de Montreal \\ 
  $^*$ indicates equal contribution\\
  \texttt{ljing@mit.edu, gulcehrc@iro.umontreal.ca} \\
}

\begin{document}
\maketitle
\begin{abstract}
We present a novel recurrent neural network (RNN) based model that combines the remembering ability of unitary RNNs with the ability of gated RNNs to effectively forget redundant/irrelevant information in its memory.  We achieve this by extending unitary RNNs with a gating mechanism. Our model is able to outperform LSTMs, GRUs and Unitary RNNs on several long-term dependency benchmark tasks. We empirically both show the orthogonal/unitary RNNs lack the ability to forget and also the ability of GORU to simultaneously remember long term dependencies while forgetting irrelevant information. This plays an important role in recurrent neural networks. We provide competitive results along with an analysis of our model on many natural sequential tasks including the bAbI Question Answering, TIMIT speech spectrum prediction, Penn TreeBank, and synthetic tasks that involve long-term dependencies such as algorithmic, parenthesis, denoising and copying tasks.
\end{abstract}

\section{Introduction}



Recurrent Neural Networks with gating units --- such as Long Short Term Memory (LSTMs) \cite{hochreiter1997long,felix2001long} and Gated Recurrent Units (GRUs) \cite{cho2014learning} --- have led to rapid progress in different areas of machine learning such as language modeling \cite{graves2014neural}, neural machine translation \cite{cho2014learning,sutskever2014sequence}, and speech recognition \cite{chan2015listen,chorowski2015attention}. These works have proven the importance of gating units for Recurrent Neural Networks.

The main advantage of using these gated units in RNNs is primarily due to the ease of optimization of the models using them and to reduce the learning degeneracies such as vanishing gradients that can cripple conventional RNNs \cite{pascanu2013difficulty}. Most importantly, by designing special gates, it is easier to impose a particular behavior on the model, such as creating shortcut connections through time by using input and forget gates in LSTMs and resetting the memory via the reset gate of a GRU. This feature also brings modularity to the neural network design that seems to make training of those models easier. Gated RNNs are also empirically shown to achieve better results for a wide variety of real-world tasks.


Recently, using unitary and orthogonal matrices (instead of general matrices) in RNNs \cite{arjovsky2015unitary,jing2016tunable,henaff2016orthogonal} have attracted an increasing amount of attention in the machine learning community. This trend was following the demonstration that these matrices can be effective in solving tasks involving long-term dependencies and gradients vanishing/exploding \cite{bengio1994learning,hochreiter1991untersuchungen} problem. Thus a unitary/orthogonal RNN can capture long term dependencies more effectively in sequential data than a conventional RNN or LSTM. As a result, this type of model has been shown to perform well on tasks that would require rote memorization\cite{hochreiter1991untersuchungen} and simple reasoning, such as the copy task\cite{hochreiter1997long} and the sequential MNIST\cite{le2015simple}. Those models can just be viewed as an extension to vanilla RNNs\cite{jordan1986serial} that replaces the transition matrices with either unitary or orthogonal matrices.

%
%


In this paper, we refer the ability of a model to omit parts of the input sequence that contain redundant information and to filter out the noise input in general as the means of a {\bf forgetting} mechanism. Previously \cite{gers1999learning} have shown the importance of the forgetting mechanism for LSTM networks and with very similar motivations, we discuss the utilization of a forgetting mechanism for RNNs with orthogonal transitions. The importance of forgetting for those networks is mainly due to that unitary/orthogonal RNNs can backpropagate the gradients without vanishing through time, and it is very easy for them to just have an output that depends on equal amounts of all the elements of the whole input sequence.  From this perspective, learning to forget can be difficult with unitary/orthogonal RNNs and they can clog up the memory with useless information. However, most real-world applications and natural tasks require the model to filter out irrelevant or redundant information from the input sequence. We argue that difficulties of forgetting can cause unitary and orthogonal RNNs to perform badly on many realistic tasks, and demonstrate this empirically with a toy task.  
%
%
%


We propose a new architecture, the Gated Orthogonal Recurrent Unit (GORU), which combines the advantages of the above two frameworks, namely (i) the ability to capture long term dependencies by using orthogonal matrices and (ii) the ability to ``forget'' by using a GRU structure. We demonstrate that GORU is able to learn long term dependencies effectively, even in complicated datasets which require a forgetting ability. In this work, we focus on implementation of orthogonal transition matrices which is just a subset of the unitary matrices. 

GORU outperforms a recent variation of unitary RNN called EURNN~\cite{jing2016tunable} on language modeling, denoising, parenthesis and the question answering tasks. We show that the unitary RNN fails catastrophically on a denoising task which requires the model to forget. On question answering, speech spectrum prediction, algorithmic, parenthesis and the denoising tasks, GORU achieves better accuracy on the test set over all other models that we compare against. We have attempted to use gates on the unitary matrices with complex numbers, but we encountered some training challenges of training gating mechanisms, thus we have decided to just to focus on orthogonal matrices for this paper.

%

\section{Background}

Given an input sequence $\x_t\in\mathbb{R}^{d_x}$, $t\in\{1, 2, \cdots, T\}$, a vanilla RNN defines a sequence of hidden states $\h_t\in\mathbb{R}^{d_h}$ updated at each time step according to the rule
\begin{equation}
\h_t = \phi(\W_h \h_{t-1} + \W_x \x_t + \b),
\end{equation}
where $\W_h\in \mathbb{R}^{d_h\times d_h}$, $\W_x\in \mathbb{R}^{d_x\times d_h}$ and $\b\in \mathbb{R}^{d_h}$ are model parameters and $\phi$ is a nonlinear activation function.
 RNNs have proven to be effective for solving sequential tasks due to their flexibility to . However, a well-known problem called gradient vanishing and gradient explosion has prevented RNNs from efficiently learning long-term dependencies \cite{bengio1994learning}. Several approaches have been developed to solve this problem, with LSTMs and GRUs being the most successful and widely used.

\subsection{Gated Recurrent Unit}

\begin{figure}[h!]
\centering
\includegraphics[height=1.2in]{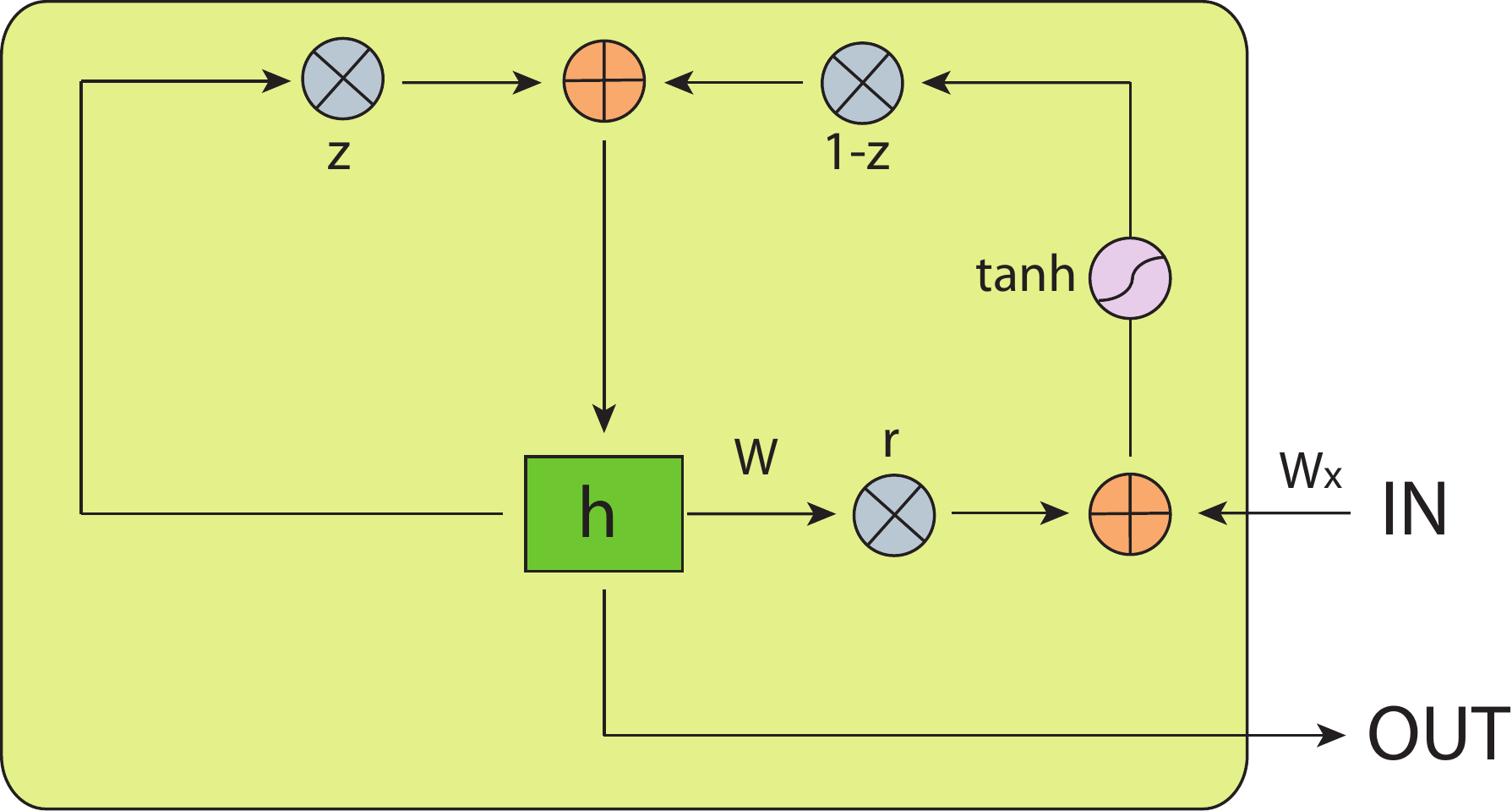}
\caption{Illustration of GRU. $r$ and $z$ are reset and update gates. $h$ is the hidden state.}
\label{fig:gru}
\end{figure}

A big step forward from LSTM is the Gated Recurrent Unit (GRU), proposed by
Cho et al, \cite{cho2014properties}, which removed the extra memory state in LSTM. Specifically, the hidden state $\h_t$ in a GRU is updated as follows:
\begin{eqnarray}
\h_t&=&\z_t \odot \h_{t-1} + (1 - \z_t) \odot \mathrm{tanh}(\W_x \x_t +\nonumber \\ 
&& \r_t \odot\W_h \h_{t-1} + \b_h)\\
\z_t &=& \mathrm{sigmoid}(\W_{z}[\h_{t-1}, \x_{t}] + \b_{z}),\\
\r_t &=& \mathrm{sigmoid}(\W_{r}[\h_{t-1}, \x_{t}] + \b_{r}),
\end{eqnarray}
where $\W_{\{z, r\}}\in \mathbb{R}^{(d_h + d_x)\times d_h}$, $\W_x\in \mathbb{R}^{d_x\times d_h}$, $\W_h\in \mathbb{R}^{d_h\times d_h}$ and $\b_{\{z, r, h\}}\in \mathbb{R}^{d_h}$. Figure \ref{fig:gru} demonstrated the architecture of GRU model.

Although LSTMs and GRUs were proposed to solve the exploding and vanishing gradient problem \cite{hochreiter1991untersuchungen,bengio1994learning} they can in practice still suffer from this issue for long-term tasks. As a result, gradient clipping \cite{pascanu2013difficulty} is usually required in the training process, although that only addresses the gradient explosion. 

\subsection{Unitary and Orthogonal Matrix RNN}
A complex-valued matrix $\U$ is unitary when it satisfies $\U\U^{*T} = \I$. A matrix $\U$ is orthogonal if it is both unitary and real-valued. Therefore, any vector $\x$ that multiplies a unitary or an orthogonal matrix satisfies:
\begin{equation}
||\U\x|| = ||\x|| 
\end{equation} 
Thanks to this property, a unitary/orthogonal matrix is able to preserve the norm of vector flows through it and thus allow for the gradient to propagate through longer time steps. Recent papers from Arjovsky et al \cite{arjovsky2015unitary,henaff2016orthogonal} pointed out that unitary/orthogonal matrices can effectively prevent the gradient vanishing/explosion problem in conventional RNNs. After this work, several other unitary/orthogonal RNN models have been proposed \cite{jing2016tunable,wisdom2016full,hyland2016learning,mhammedi2016efficient}, all showing promising abilities in capturing long term dependencies in data.

A unitary/orthogonal matrix RNN is simply defined as replacing the hidden to hidden matrices in a vanilla RNN by unitary/orthogonal matrices:
\begin{equation}
\h_t = \sigma(\U \h_{t-1} + \W_x \x_t, \b) .
\end{equation}
For unitary matrices, the nonlinear activation function $\sigma$ needs to handle complex-valued inputs.
In this paper, we use the generalizations of the popular real-valued activation function $\mathrm{ReLU}(x)=\max(0,x)$ known as
\begin{equation}
\mathrm{modReLU}(z_i, b_i)\equiv\frac{z_i}{|z_i|}\mathrm{ReLU}(|z_i| + b_i),
\end{equation}
where $b$ is a bias vector.
This variant was found to perform effectively on a suite of benchmarks in \cite{arjovsky2015unitary,jing2016tunable}. Even though $modReLU$ was developed for complex value models, it turns out this activation function fits unitary/orthogonal matrices best. 




\section{The Gated Orthogonal Recurrent Unit}

\subsection{The Problem of Forgetting in Orthogonal RNNs}

First, we argue for the advantage of an RNN which can forget some of its past inputs. This is desirable because we seek a state representation which can capture the most important elements of the past sequence, and can throw away irrelevant details or noise. This ability becomes particularly critical when the dimensionality of the RNN state is smaller than the product of the sequence length with the input dimension; i.e., when some form of compression is necessary. For this compression to be most useful for further processing, it is likely that it requires a non-linear combination of the past input values, allowing the network to forget and ignore unnecessary elements from the past.

Now consider an RNN whose state is obtained as a sequence of orthogonal transformations, with each transformation being a function of the input at a given time step. Let us focus on class of orthogonal transformations that are basically rotation for the simplicity of our analysis, which (non-commutativity aside) are analogous to addition in the space of angles. When we compose several orthogonal operators, we just add more angles together. So we forget in the mild sense that we get in the state a combination of several rotations (like adding the angles) and we lose track of exactly which individual rotations were applied. The advantage is that, in the space of angles, the derivative of the final angle to any of the individually added angle is 1, so there is no vanishing gradient. However, we cannot have complete forgetting, e.g., making the new state independent of the past inputs (or of some of them which we wish to forget): for a new rotation to cancel an old rotation, one would need the new rotation to "know" about the old rotation to cancel, i.e., it would need to be a function of the old rotation. But this is not what happens, because each new rotation is chosen before looking at the current state. Instead, in a regular RNN, the state update depends in a non-linear way on the past state, so that for example when a particular value of the state is reached, it can be reset to 0. This would not be possible with just the composition of orthogonal transformations. These considerations motivate an architecture in which we combine orthogonal or unitary transformations with non-linearities which can be trained to forget when and where it is appropriate.

\subsection{GORU Architecture}

\begin{figure}[h!]
\centering
\includegraphics[height=1.2in]{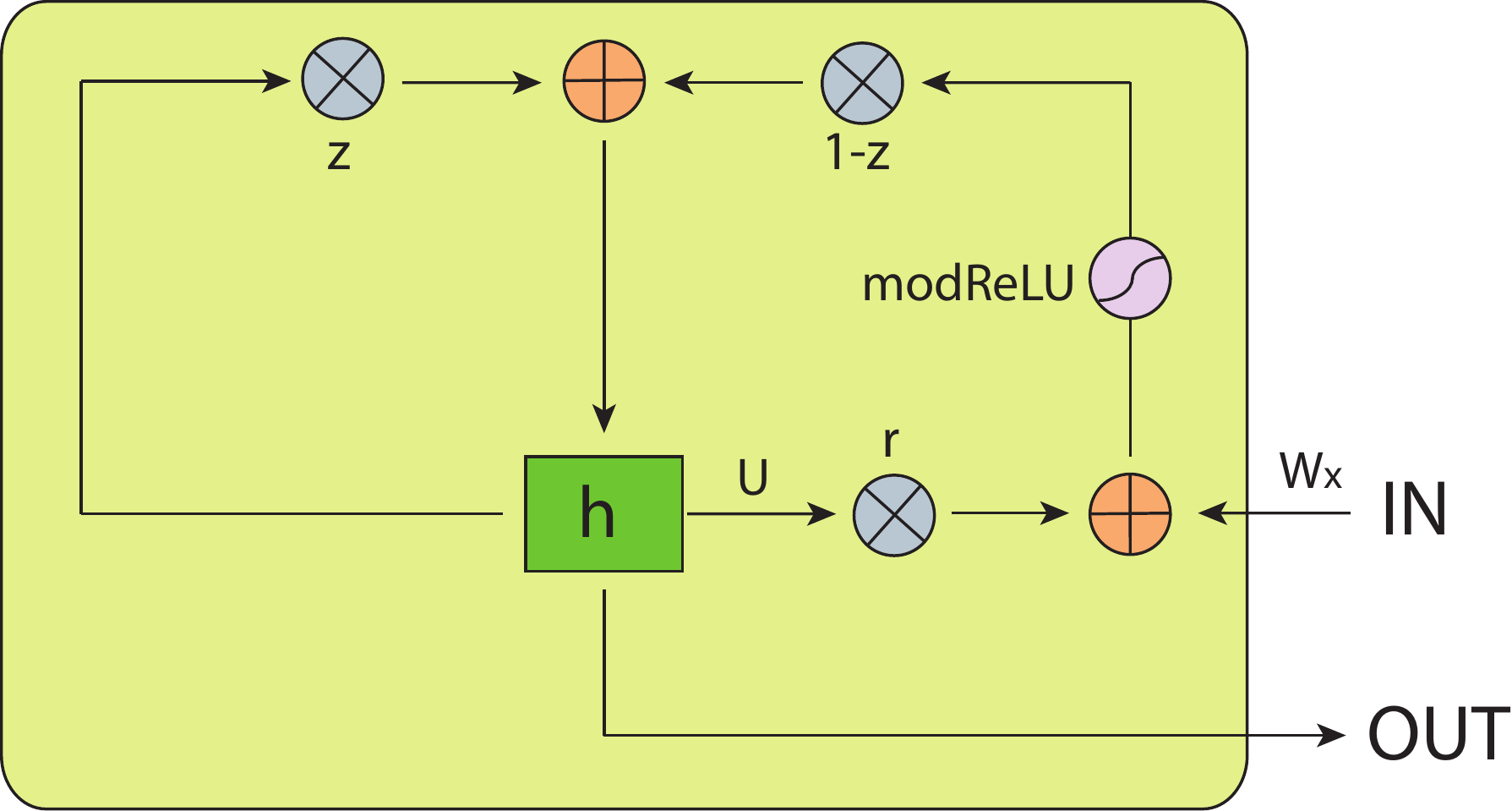}
\caption{Illustration of GORU. $h$ is the hidden state. For GORU, $r$ and $z$ are reset and update gates. It uses modReLU activation function instead of tanh.}
\label{fig:goru}
\end{figure}

This section introduces the Gated Orthogonal Recurrent Unit (GORU). In our architecture, we change the hidden state loop matrix into an orthogonal matrix and change the respective activation function to modReLU:
\begin{eqnarray}
\h_t &=& \z_t \odot \h_{t-1} + (1 - \z_t) \odot \mathrm{modReLU}(\W_x\x_t + \nonumber \\
&&\r_t \odot( \U\h_{t-1})+ \b_{h}),\\
\z_t &=& \mathrm{sigmoid}(\W_{z}\h_{t-1} + \W_{z,x}\x_{t} + \b_{z}),\\
\r_t &=& \mathrm{sigmoid}(\W_{r}\h_{t-1} + \W_{r,x}\x_{t} + \b_{r}),
\end{eqnarray}
where $\sigma$ is a suitable nonlinear activation function and 
$\W_z, \W_r \in \mathbb{R}^{d_h\times d_h}$, $\b_z, \b_r, \b_h \in \mathbb{R}^{d_h}$, $\W_{z, x}, \W_{r, x}, \W_x \in \mathbb{R}^{d_x\times d_h}$. 
$\r_t$ and $\z_t$ are the reset and update gates, respectively. $\U \in\mathbb{R}^{d_h\times d_h}$ is kept orthogonal. In fact, we have only modified the main loop that absorbs new information to orthogonal while leaving the gates unchanged compared to the GRU. Figure \ref{fig:goru} demonstrated the architecture of GORU model.

\begin{figure}[h!]
\centering
\includegraphics[height=1.2in]{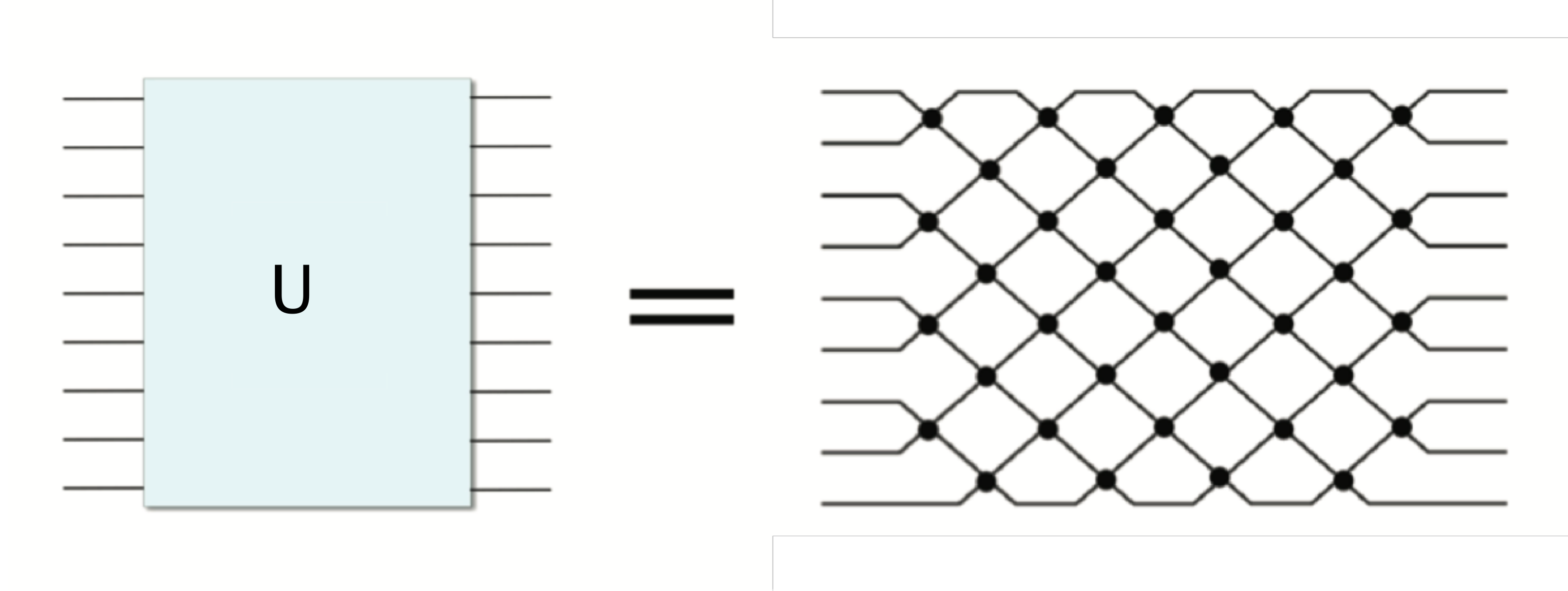}
\caption{Orthogonal matrix parametrization by 2-by-2 rotation matrices. Each row represents one neuron in the hidden state. Each junction represents a 2-by-2 rotation matrix on corresponding two neurons.}
\label{fig:unitary}
\end{figure}

We enforce matrix $\U$ to be orthogonal by using parametrization method purposed in \cite{jing2016tunable}. $\U$ is decomposed into a sequence of 2-by-2 rotation matrices as shown in Figure \ref{fig:unitary}. Each 2-by-2 rotation contains one trainable rotation parameter.

The update gates of the GORU help the model to filter out irrelevant or noise information coming from the input. It can be thought of as acting like a low-pass filter. The orthogonal transition matrices help the model to prevent the gradients to vanish through time. However, the ways an orthogonal transformation can interact with the hidden state of an RNN is limited to reflections and rotations. The reset gate enables the model to rescale the magnitude of the hidden state activations ($\h_t$).

\section{Experiments}
We compare GORU with unitary RNNs (using the EURNN parameterization purposed by Jing et al.\cite{jing2016tunable}) and two other well-known gatedRNNs (LSTMs and GRUs). Previous research on unitary RNNs has mainly focused on memorization tasks; in contrast, we focus on more realistic noisy tasks, which require the model to discard parts of the input sequence to be able to use its capacity efficiently. 

GORU is implemented in Tensorflow, available from \url{https://github.com/jingli9111/GORU-tensorflow} 

\subsection{Copying Memory Task}

The first task we consider is the well known {\it Copying Memory Task}. The copying task is a synthetic task that is commonly used to test the network's ability to remember information seen $T$ time steps earlier. 

Specifically, the task is defined as follows.
An alphabet consists of symbols  $\{a_i\}, i \in \{0, 1, \cdots, n-1, n, n+1\}$, the first $n$ of which represent data, and the remaining two representing ``blank'' and ``marker", respectively. Here we choose $n=8$.
The input sequence contains 10 data steps, followed by ``blank''. The RNN model is supposed to output ``blank'' and give the original sequence once it sees the ``marker''. Note that each instance has a different location for these 10 elements of data.

In this experiment, we use RMSProp optimization with a learning rate of 0.001 and a decay rate of 0.9 for all models. The batch size is set to 128. All these models have roughly same number of hidden to hidden parameters, despite not having similar neuron layer sizes.
\begin{figure}[h!]
\centering
\includegraphics[width=1.0\linewidth]{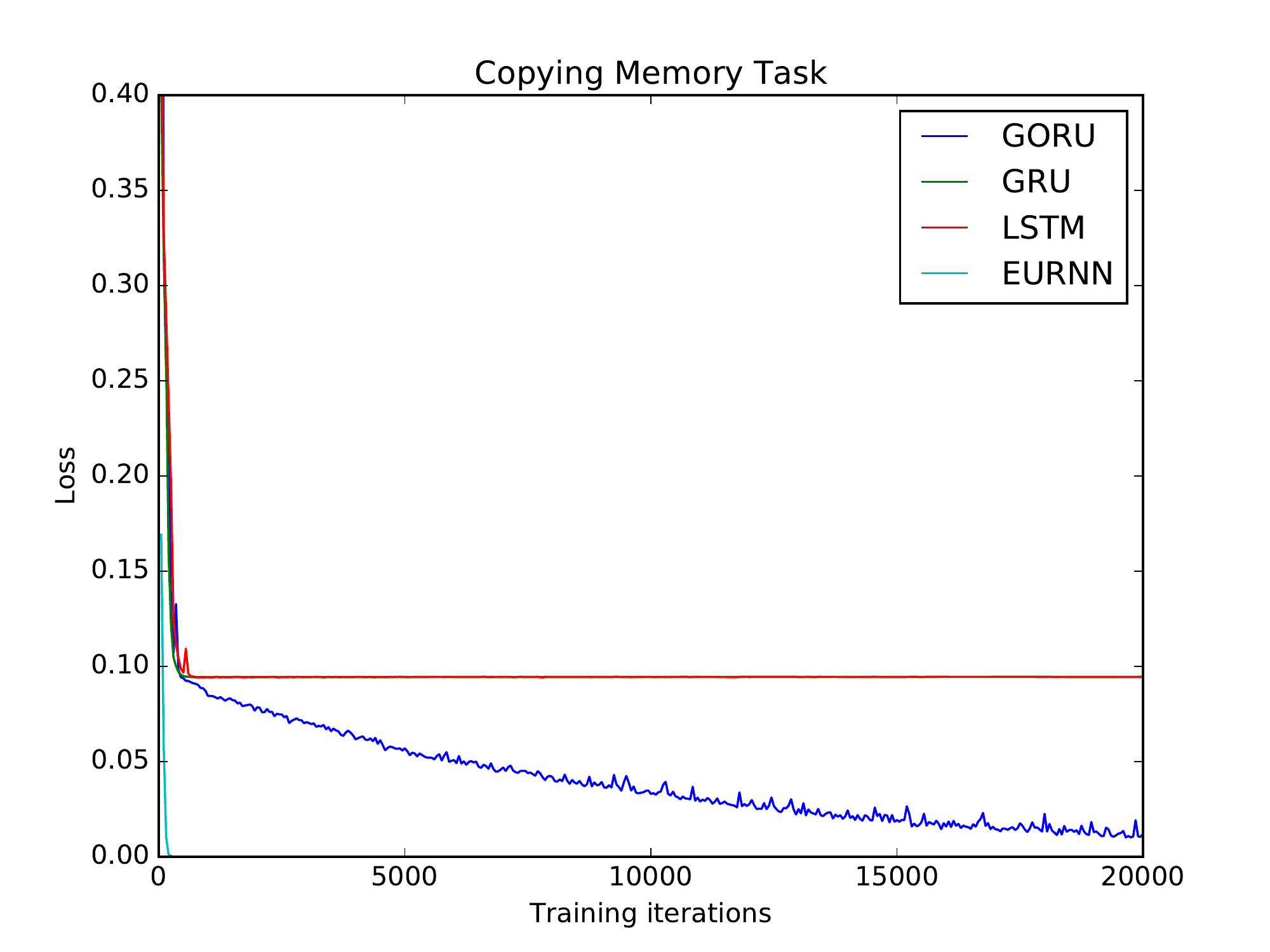}
\caption{Copying Memory Task with delay time $T=200$ on GORU, GRU, LSTM and EURNN. Hidden state sizes are set to 128, 100, 90, 512, respectively to match total number of hidden to hidden parameters. GORU is the only gated-system to successfully solve this task while the GRU and LSTM both get stuck at the baseline. EURNN is seen to converges within hundreds of iterations. }
\label{fig:copying}
\end{figure}

This task only requires the model to efficiently overcome the gradient vanishing/explosion problem and does not require a forgetting ability. The EURNN performs perfectly and goes through to the baseline in no time --- as previously seen.
The GORU is the only gated-system to successfully solve this task while the GRU and LSTM get stuck at the baseline as shown in Figure \ref{fig:copying}.

\subsection{Denoise Task}

We evaluate the forgetting ability of each RNN architecture on a synthetic {\it "denoise" task}. A list of data points are located randomly in a long noisy sequence. The RNN model is supposed to filter out the useless part ("noise") and output the remaining sequential labels.

Similarly to the labels of copying memory task above,
an alphabet consists of symbols  $\{a_i\}, i \in \{0, 1, \cdots, n-1, n, n+1\}$, the first $n$ of which represent data, and the remaining two represent ``noise'' and ``marker", respectively.
The input sequence contains 10 randomly located data steps and the rest are filled by ``noise''. The RNN model is supposed to output those 10 data in a sequence after it sees the ``marker''.
Just as in the previous experiment, we use RMSProp optimization algorithm with a learning rate of 0.01 and a decay rate of 0.9 for all models. The batch size is set to 128. All these models have roughly same number of hidden to hidden parameters.

\begin{figure}[h!]
\centering
\includegraphics[width=1.0\linewidth]{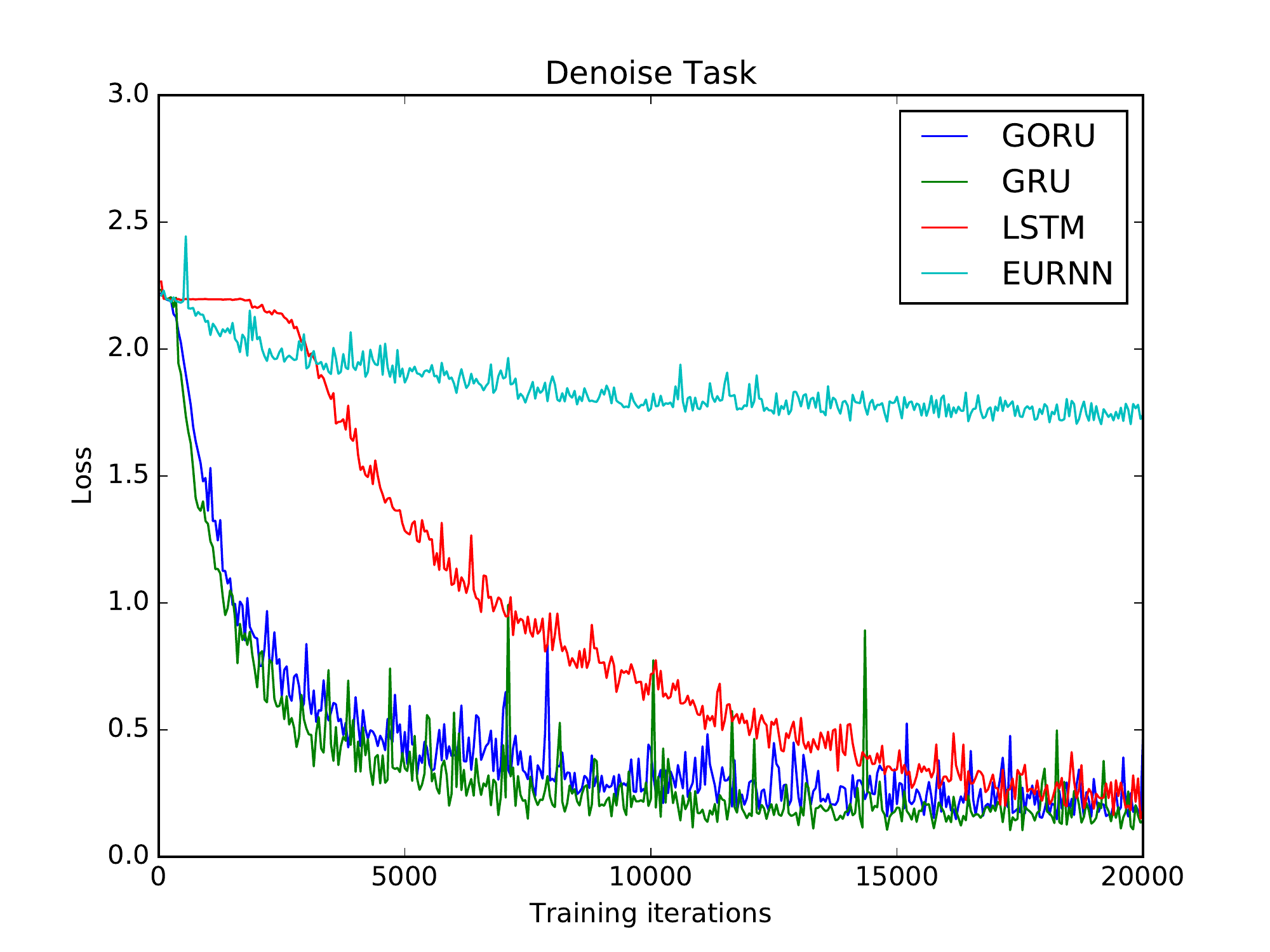}
\caption{Denoise Task with sequence length $T=200$ on GORU, GRU, LSTM and EURNN.  Hidden state sizes are set to 128, 100, 90, 512, respectively to match total number of hidden to hidden parameters. EURNN get stuck at the baseline because of lacking forgetting mechanism, while GORU and GRU successfully solve the task.}
\label{fig:denoise}
\end{figure}

This task requires both the ability of learning long dependencies but also the ability to forget the noisy input. 
The GORU and GRU both are able to successfully outperform LSTM in terms of both learning speed and final performances as shown in Figure \ref{fig:denoise}. EURNN, however, gets stuck at the baseline, just as we intuitively expected.


\subsection{Parenthesis Task}

The parenthesis task\cite{foerster2016intelligible} requires the RNN model to count the number of each type of unmatched parentheses at each time step, given that there are 10 types of parentheses. The input data contains 10 pairs of different parenthesis types --- e.g. '(, [, \{, <, ), ], \}, >, ...' --- mixed with random noise/characters between them. The neural network outputs how many unmatched parentheses there are. For instance, given '(((', the neural network would output '123'.Note that there are never more than 10 unmatched parentheses in any category. 

In our experiment, the total input length is set to 200. We used batch size 128 and RMSProp Optimizer with a learning rate 0.001, decay rate 0.9 on all models. Hidden state sizes are set to match their total numbers of hidden to hidden parameters.

\begin{figure}[h!]
\centering
\includegraphics[width=1.0\linewidth]{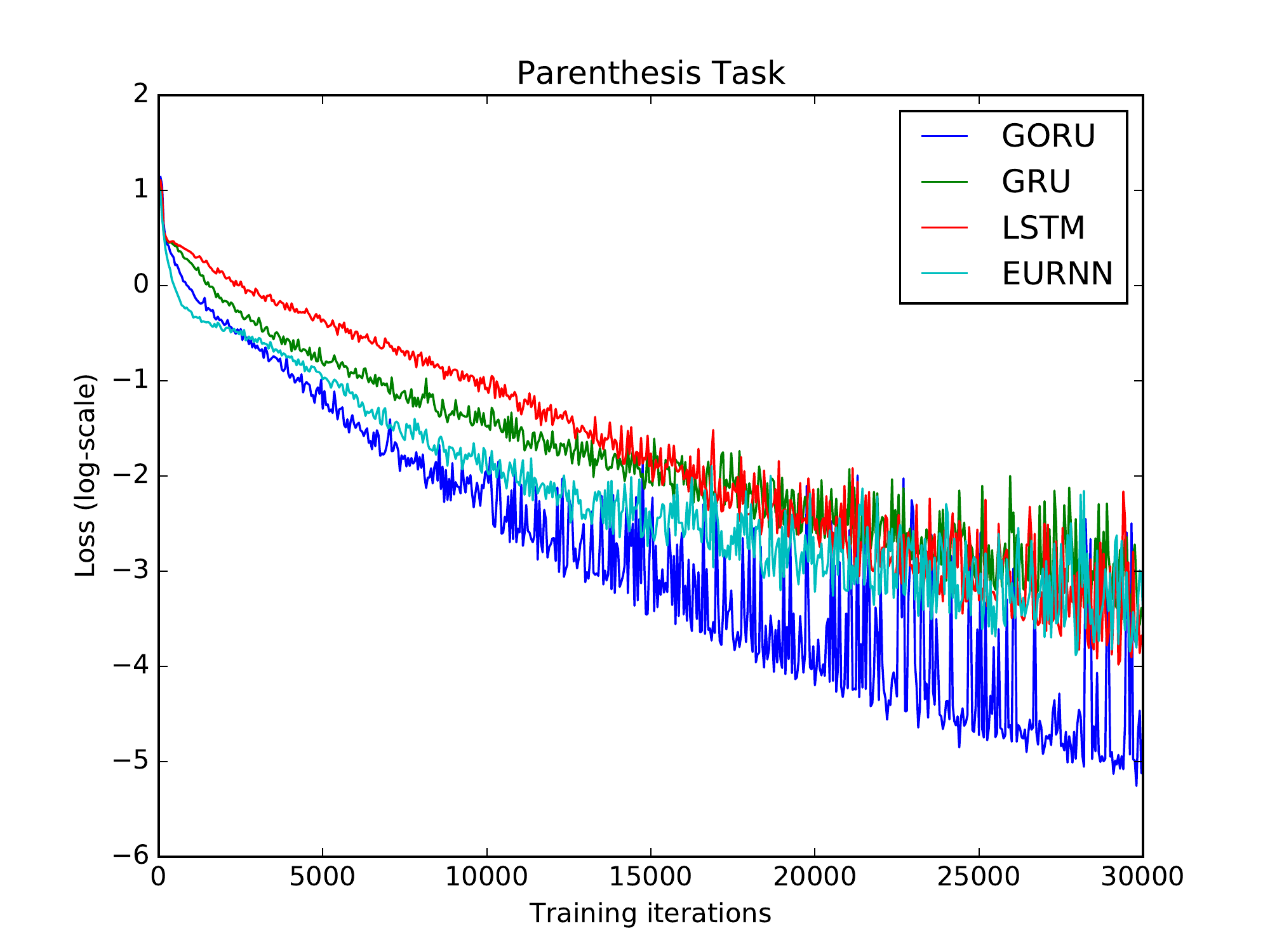}
\caption{Parenthesis tasks with total sequence length $T=200$ on GORU, GRU, LSTM and EURNN. Hidden state sizes are set to 128, 100, 90, 512, respectively to match total number of hidden to hidden parameters. Loss is shown in log scale. GORU is able to outperform all other models in both learning speed and final performance.}
\label{fig:paren}
\end{figure}

This task requires learning long-term dependencies and forgetting of the noisy data.
The GORU is able to successfully outperform GRU, LSTM and EURNN in terms of both learning speed and final performances as shown in Figure \ref{fig:paren}.

\begin{figure}[h!]
\centering
\includegraphics[width=1.0\linewidth]{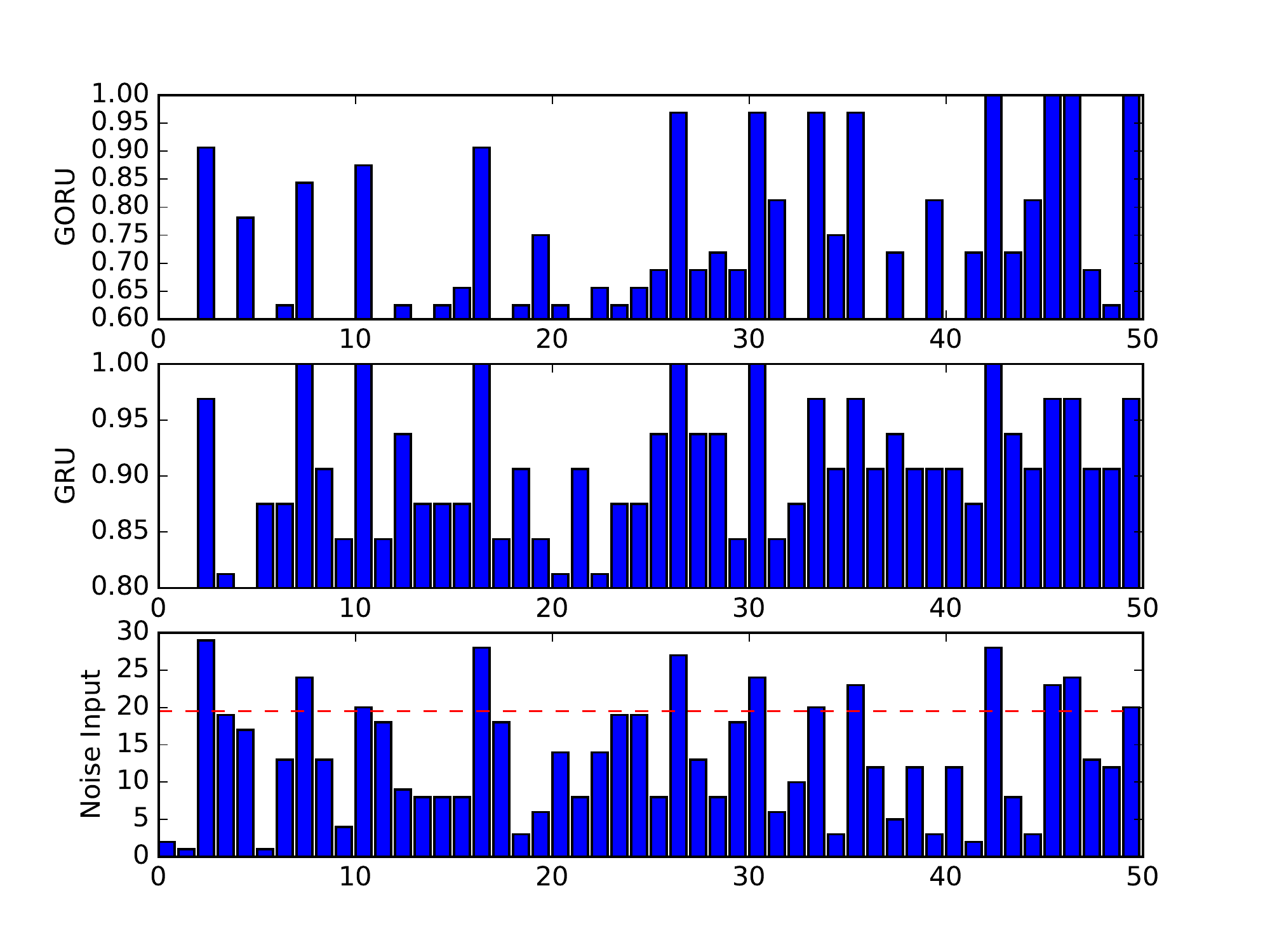}
\caption{The activations of the update gates for GORU and GRU on the parenthesis task. In those bar-charts we visualize the number of activations that are greater than 0.7 normalized by the total number of units in the gate, 
$\frac{\sum_{i=0}^{d_h}{1}_{\r_i > 0.7}}{d_h}$. 
As can be seen in the top two plots, the activations of the update gate peaks and becomes almost one when the input is noise when the magnitude of the noise input in the bottom plot is above the red bar.}
\label{fig:bar}
\end{figure}

We also analyzed the activations of the update gates for GORU and GRU. According to the histogram of activations shown in Figure \ref{fig:bar}, both models behave very similarly, and when the model receives noise as input, the activations of its update gate peaks. This behavior helps the model to forget and omit the noise input.

\subsection{Algorithmic Task}
We tested the RNN models on algorithmic task as described in \cite{li2015gated}. The model is fed with random graph as an input sequence and required to output the shortest path at the end of the sequence. We have used the exact same setup and use the data provided as in \cite{li2015gated}. 

We used batch size 50 and hidden size 128 for all models. The RNNs are trained with RMSProp optimizer with a learning rate of 0.001 and decay rate of 0.9.

We summarized the test set results in Table \ref{tbl:algo_tasks}. We found that the GORU performs averagely better than GRU/LSTM and EURNN.

\begin{table}[h!]
\centering
\begin{tabular}{cc}
\hline
Model & Accuracy \\
\hline
EURNN & $66.3\pm3.2$ \\
LSTM & $60.9\pm4.7$ \\
GRU & $74.1\pm4.8$ \\
GORU & $\textbf{75.0}\pm5.3$ \\
\hline
\end{tabular}
\caption{Algorithmic Test on GORU, GRU, LSTM and EURNN.}
\label{tbl:algo_tasks}
\end{table}

\subsection{bAbI: Episodic Question Answering}
We tested the ability of our RNN models on a word-level episodic question answering task. The bAbI dataset \cite{weston2015towards} examines RNN's ability to understand language and perform basic logical reasoning. Each training example is a set of statements, that are logically related in some fashion. For instance, one training example consists of these three statements and a question: {\it Mary went to the bathroom. John moved to the Hallway. Mary traveled to the office. Where is Mary?} Answer: o{\it ffice}.

There are twenty different types of questions that can be asked --- some requiring deduction between lines, some requiring association. The bAbI dataset is useful because it contains a small sized vocabulary, short sentences and requires one word answers for each story. Thus, it is a good benchmarking test because the word mapping layers are not the dominant sources of parameters.  

We test each task with a uni-directional RNN without any attention mechanism. In detail, we word-embed and then feed one RNN the sequence of statements. Another RNN is fed the word-embedded question. Then, we concatenate the outputs of the two RNN's into a single input for a third RNN that then outputs the correct word. 

We summarized the test set results as follows in Table \ref{tbl:babi}. We found that the GORU performs averagely better than GRU/LSTM and EURNN. We also show the big gains over EURNN by introducing the gates.

\begin{table*}[!htb]
\centering
\begin{tabular}{lccccc}
\hline
Task           & GORU & GRU & LSTM & EURNN &  baseline \cite{weston2015towards}\\
\hline
1 - Single Supporting Fact	&	45.8	&	49.1	&	49.3	&	47.2	&	50	\\
2 - Two Supporting Facts	& 	39.5	&	38.5	&	32.3	&	24.3	&	20	\\
3 - Three Supporting Facts	&	33.5	&	32.2	&	20.6	&	22.5	&	20	\\
4 - Two Arg. Relations	&	62.7	&	64.6	&	67.5	&	56.1		&	61	\\
5 - Three Arg. Relations	&	\textbf{87.0}	&	78.0	&	52.3	&	56.2	&	70	\\
6 - Yes/No Questions	&	53.6	&	50.5	&	49.3	&	50.5	&	48	\\
7 - Counting	&	77.7		&	79.5	&	76.9	&	71.9	&	49	\\
8 - Lists/Sets	&	75.0	&	75.5	&	76.8	&	56.5	&	45	\\
9 - Simple Negation	&	62.9	&	63.9	&	63.5	&	60.6	&	64	\\
10 - Indefinite Knowledge	&	45.4	&	44.8	&	46.0	&	42.6	&	44	\\
11 - Basic Coreference	&	69.3	&	71.2	&	71.1	&	72.1	&	72	\\
12 - Conjunction	&	69.9	&	71.6	&	71.9	&	72.7	&	74	\\
13 - Compound Coref.	&	92.7	&	94.2	&	93.8	&	92.4	&	94	\\
14 - Time Reasoning	&	37.9	&	39.2	&	34.4	&	20.0	&	27	\\
15 - Basic Deduction	&	55.2	&	57.4	&	20.9	&	25.0	&	21	\\
16 - Basic Induction	&	44.0	&	45.9	&	45.9	&	43.3	&	23	\\
17 - Positional Reasoning	&	\textbf{59.6}	&	50.5	&	51.6	&	51.2	&	51	\\
18 - Size Reasoning	&	90.5	&	89.9	&	91.8	&	89.7		&	52	\\
19 - Path Finding	&	8.9		&	9.6		&	8.2		&	9.0	&	8.0	\\
20 - Agent's Motivations	&	97.7	&	97.7	&	96.5	&	93.3	&	91		\\
\hline
Mean Performance	&	\textbf{60.4}	&	58.2	&	56.0	&	52.9	&	49.2		\\
\hline
\end{tabular}
\caption{Question Answering task on bAbI dataset. Test accuracy (\%) on GORU, GRU, LSTM and EURNN. All models are performed on one-way RNN without extra memory or attention mechanism. GORU achieves highest average accuracy.}
\label{tbl:babi}
\end{table*}

\subsection{Language Modeling: Character-level Prediction}

We test each RNN on character-level language modeling. The RNN is fed by one character each step from a real context and supposed to output the prediction for the next character. 
We used the Penn Treebank corpus \cite{marcus1993building}. 

We use RMSProp with minibatch size of 32 and a learning rate of 0.001. Each training sequence is unfolded into 50 time steps. Similar to most work in language modeling, at the end of each sequence, the hidden state is saved and used to initialize the hidden state for the next sequence. This allows the neural network to give consistent predictions even at the beginning of a sequence.

We show the final test performance in Table \ref{tbl:ptb} by comparing their performance in terms of bits-per-character. GORU is performing comparable to LSTM and GRU in our experiments and it performs significantly better than EURNN. We have also done an ablation study with disabling reset and update gates.
Since most of the relevant information for character-level prediction can be obtained by only using the recent rather than distant past \cite{karpathy2015visualizing}, the core of the character-prediction challenge does not involve the main strength of EURNN.  


\begin{table}[h!]
\centering
\begin{tabular}{lcc}
\hline
Model           & bpc & \# Units\\
\hline
LSTM          & 1.596 & 350 \\
GRU         & 1.601  & 415 \\
EURNN       & 1.715  & 2048 \\
GORU       & 1.623  & 512 \\
GORU (w/o reset gate)       & 1.722  & 512 \\
GORU (w/o update gate)       & 1.754  & 512 \\
\hline
\end{tabular}
\caption{Penn Treebank character-level modeling Test on GORU, GRU, LSTM and EURNN. We only use single layer models. We choose the size of the models to match the number of parameters. GORU is able to outperform EURNN. We also tested the performance of restricted GORU which shows the necessity of both reset and update gates.}
\label{tbl:ptb}
\end{table}

\subsection{Speech Spectrum Prediction}
We tested the ability of our RNN models on real-world speech spectrum prediction task in short-time Fourier transform (STFT)\cite{wisdom2016full,jing2016tunable}.
We used TIMIT dataset sampled in 8 kHz. The audio .wav file is initially divided into different time frames and then Fourier transformed into the frequency domain and finally normalized for training/testing. In our STFT operation we uses a Hann analysis window of 256 samples (32 milliseconds) and a window hop of 128 samples (16 milliseconds). In this task, the RNNs are required to predict th log-magnitude of the STFT frame at time t + 1, given all the log-magnitudes of STFT frames up to time t.

We used a training set with 2400 utterances, a validation set of 600 utterances and 1000 utterances for test. We trained all RNNs for with the same batch size 32 using Adam optimization with a learning rate of 0.001.

We found GORU significantly outperforms all other models with same hidden size as shown in Table \ref{tbl:timit}.

\begin{table}[h!]
\centering
\begin{tabular}{lccc}
\hline
Model       & \#parameters  & MSE(validation) &  MSE(test) \\
\hline
LSTM    & 98k & 58.8 & 57.5 \\
GRU      & 72k & 58.9 & 57.3  \\
EURNN   & 41k & 51.8 & 51.9 \\
GORU    & 59k & \textbf{45.4}  & \textbf{47.6}  \\
\hline
\end{tabular}
\caption{Speech Spectrum prediction Test on GORU, GRU, LSTM and EURNN. We only use single layer models with same hidden size 128.}
\label{tbl:timit}
\end{table}

\section{Conclusion}

  We have built a novel RNN that brings the benefits of orthogonal matrices to gated architectures: the Gated Orthogonal Recurrent Units (GORU). By replacing the hidden to hidden matrix in the reseting path of the GRU to be an orthogonal matrix, and replacing the non-linear activation to a modReLU, GORU gains the advantage of unitary/orthogonal RNNs since the gradient can pass through long time steps without exploding.
   Our empirical results showed that GORU is the only model we found that could solve both the synthetic copying task and the denoise task. Moreover, GORU is able to outperform GRU and LSTM in several benchmark tasks.  
   
These results suggest that the GORU is the first step in bringing an explicit forgetting mechanism to the class of unitary/orthogonal RNNs. Our method demonstrates how to incorporate orthogonal matrices into a variety of neural network architectures, and we are excited to open the gate the next level of neural networks.

\section*{Acknowledgment}
This work was partially supported by the Army Research
Office through the Institute for Soldier Nanotechnologies
under contract W911NF-13-D0001, the National Science
Foundation under Grant No. CCF-1640012, and by the Semiconductor Research Corporation under
Grant No. 2016-EP-2693-B.

\bibliography{citations}
\bibliographystyle{aaai}

\end{document}